\documentclass{article}

\usepackage{microtype}
\usepackage{graphicx}
\usepackage{subfigure}
\usepackage{booktabs} 

\usepackage{hyperref}

\usepackage[accepted]{icml2023}

\usepackage{amsmath}
\usepackage{amssymb}
\usepackage{mathtools}
\usepackage{amsthm}

\usepackage[capitalize,noabbrev]{cleveref}

\theoremstyle{plain}

\theoremstyle{definition}

\theoremstyle{remark}

\usepackage[textsize=tiny]{todonotes}

\usepackage{ctable}
\usepackage{multirow}

\icmltitlerunning{Gradient Scaling on Deep Spiking Neural Networks with Spike-Dependent Local Information}

\begin{document}

\twocolumn[
\icmltitle{Gradient Scaling on Deep Spiking Neural Networks\\
with Spike-Dependent Local Information}



\icmlsetsymbol{equal}{*}

\begin{icmlauthorlist}
\icmlauthor{Seongsik Park}{kist}
\icmlauthor{Jeonghee Jo}{kist}
\icmlauthor{Jongkil Park}{kist}
\icmlauthor{Yeonjoo Jeong}{kist}
\icmlauthor{Jaewook Kim}{kist}
\icmlauthor{Suyoun Lee}{kist}
\icmlauthor{Joon Young Kwak}{kist}
\icmlauthor{Inho Kim}{kist}
\icmlauthor{Jong-Keuk Park}{kist}
\icmlauthor{Kyeong Seok Lee}{kist}
\icmlauthor{Gye Weon Hwang}{kist}
\icmlauthor{Hyun Jae Jang}{kist}
\end{icmlauthorlist}

\icmlaffiliation{kist}{Center for Neuromorphic Engineering, Korea Institute of Science and Technology, Seoul, Korea}

\icmlcorrespondingauthor{Seongsik Park}{seong.sik.park@kist.re.kr}

\icmlkeywords{Machine Learning, ICML}

\vskip 0.3in
]



\printAffiliationsAndNotice{}  

\begin{abstract}

Deep spiking neural networks (SNNs) are promising neural networks for their model capacity from deep neural network architecture and energy efficiency from SNNs' operations.
To train deep SNNs, recently, spatio-temporal backpropagation (STBP) with surrogate gradient was proposed.
Although deep SNNs have been successfully trained with STBP, they cannot fully utilize spike information.
In this work, we proposed gradient scaling with local spike information, which is the relation between pre- and post-synaptic spikes.
Considering the causality between spikes, we could enhance the training performance of deep SNNs.
According to our experiments, we could achieve higher accuracy with lower spikes by adopting the gradient scaling on image classification tasks, such as CIFAR10 and CIFAR100.

\end{abstract}

\section{Introduction} \label{submission}

Deep learning with deep neural networks (DNNs) have been rapidly advancing artificial intelligence (AI) technology in various fields~\cite{lecun2015deep,tan2019efficientnet}.
However, as AI technology continues to progress, it demands more energy and computing resources, raising concerns about sustainable development and application.
Spiking neural networks (SNNs) have received considerable attention as a solution to this problem.
SNNs, which have been considered third-generation artificial neural networks, enable event-based computing, resulting in sparse operations compared to DNNs~\cite{maass1997networks}.
Furthermore, SNNs hold great importance as the basis for neuromorphic computing, which imitates the operations of the human brain for its exceptional energy efficiency~\cite{davies2018loihi, roy2019towards}.

Deep SNNs have been actively studied to combine the features of both DNNs and SNNs: the model capacity of the former and the energy efficiency of the latter. 
Deep SNNs have a similar synaptic topology to DNNs, with interconnected spiking neurons. 
While deep SNNs can leverage the advantages of both DNNs and SNNs, they face challenges in training.
As an indirect training method for deep SNNs, DNN-to-SNN conversion has been proposed.
Although this approach has enabled the implementation of various deep SNN models~\cite{park2019fast, kim2020spiking, park2020t2fsnn, li2021free, bu2023optimal}, it has introduced issues, such as long inference latency~\cite{han2020rmp}.


Recently, to improve the training performance, a gradient-based training algorithm, which is a successful training approach for DNNs, has been applied to training deep SNNs, such as spatio-temporal backpropagation (STBP)~\cite{wu2018spatio}.
This method with surrogate gradient, which can handle the non-differentiability of spiking neurons, has proven to be effective in training deep SNNs.
Based on the successful training, gradient-based training approaches have inspired further research about improving the training performance of deep SNNs~\cite{zheng2021going, deng2022temporal, yang2022training}.
Furthermore, it has enabled the expansion of deep SNNs in various applications and algorithms, including Transformer models~\cite{zhou2022spikformer} and neural architecture search algorithms~\cite{na2022autosnn}.

Gradient-based training algorithms have allowed deep SNNs to utilize their model capacity sufficiently.
However, these algorithms cannot exploit the dynamic characteristics of SNNs as they are derived from DNNs.
Unlike DNNs, SNNs have spatio-temporal features, and spiking neurons transmit information in the form of spikes.
Thus, to maximize the training performance of deep SNNs, we proposed a training algorithm, called gradient scale, that can consider the spike dynamics in SNNs.
We were inspired by spike-timing-dependent plasticity (STDP), which is a biologically plausible training algorithm of SNNs with local spike causality~\cite{diehl2015unsupervised}.
While utilizing the training performance of gradient-based algorithms, we adjusted gradients depending on the local spike relationships, which can be defined by the causality between spikes of pre- and post-synaptic neurons.
The proposed algorithm was evaluated on ResNet architectures~\cite{he2016deep} with image classification tasks, such as CIFAR10 and CIFAR100~\cite{Krizhevsky09learningmultiple}.

\section{Related Works}
\subsection{Spiking Neural Networks}
SNNs consist of spiking neurons and synapses that connect them.
Mimicking the behavior of the brain, spiking neurons exchange information with binary spikes through synapses.
Because of the spike-based operation, SNNs have been expected to enable event-driven computing, which is a next-generation and energy-efficient computing paradigm.
Thus, SNNs are promising neural networks for energy-efficient artificial intelligence as a fundamental component of neuromorphic computing that mimics the operations of the human brain.

Although there are various types of spiking neurons, such as izhikevich, leaky integrate-and-fire (LIF), and integrate-and-fire (IF) neuron models~\cite{izhikevich2004model}, neurons commonly operate in an integrate-and-fire manner.
Spiking neurons integrate incoming information into the internal state, called membrane potential, and fire spikes whenever the potential exceeds a threshold voltage.
Due to the low complexity of computation, most deep SNNs adopt relatively simple spiking neuron models, such as IF and LIF.
Thus, in this work, we used an LIF neuron model, which is described as
\begin{equation}
u_{j}^{l}(t) = \tau u_{j}^{l}(t\textrm{-}1) + z_{j}^{l}(t) \textrm{,}
\end{equation}
where $\tau$ is a leak constant, $u_{j}^{l}(t)$ and $z_{j}^{l}(t)$ are the membrane potential and incoming information of the $j$th spiking neuron in $l$th layer at time step $t$, respectively.
The incoming information, called post-synaptic potential (PSP), is caused by pre-synaptic spikes (input spikes) as
\begin{equation} \label{eq:psp}
z_{j}^{l}(t) = \sum_{i}{w_{ij}^{l}s_{i}^{l\textrm{-}1}(t)+b_{j}^{l}} \textrm{,}
\end{equation}
where $w$ and $b$ are the synaptic weight and bias, respectively.
When the accumulated information on the membrane potential exceeds a certain threshold, spikes are generated, and the information is transmitted to adjacent neurons through synapses.
Spike generation can be expressed as
\begin{equation} \label{eq:fire}
s_{j}^{l}(t) = H(u_{j}^{l}(t)-v_{\textrm{th},j}^{l}(t)) \textrm{,}
\end{equation}
where $H$ is the Heaviside step function, and $v_{\textrm{th}}$ is a threshold voltage.
When a spike is generated, the membrane potential is reset.
There are mainly two reset methods: soft and hard reset can be stated as
\begin{equation} \label{eq:vrest}
u_{j}^{l}(t) = 
\begin{cases}
    u_{j}^{l}(t) - s_{j}^{l}(t) v_{\textrm{th},j}^{l}(t)    & \textrm{(soft)} \\
    (s_{j}^{l}(t)-1) u_{j}^{l}(t) + s_{j}^{l}(t) v_{\textrm{r},j}^{l}(t) & \textrm{(hard)} \textrm{,} \\
\end{cases}
\end{equation}
where $v_{\textrm{r}}$ is a rest potential.

%


\subsection{Training Methods of deep SNNs}

Training algorithms of deep SNNs can be categorized into two approaches: indirect and direct training.
Indirect training, which is represented by DNN-to-SNN conversion, transforms a pre-trained DNN model into deep SNN with the same topology, and the converted SNN only performs inference.
This approach has been successfully applied to various neural network architectures~\cite{sengupta2019going,han2020rmp}, applications~\cite{kim2020spiking}, and neural codings~\cite{park2019fast,zhang2019tdsnn,park2020t2fsnn}.
However, it had drawbacks, such as long inference latency, due to disregarding features of SNNs during the training of DNNs.
Certain studies have attempted to address these limitations with calibration~\cite{li2021free} and SNN-aware DNN training~\cite{bu2023optimal}, but there still remain limitations that it is challenging to directly consider the dynamics of deep SNNs.

\begin{figure}[t]
    \centering
    \includegraphics[width=1.0\linewidth]{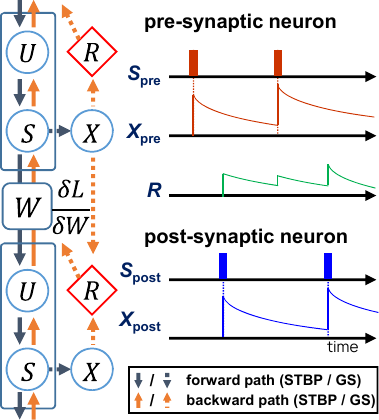}
    \caption{Spike trace and proposed gradient scaling (GS).}
    \label{fig:spike_trace}
\end{figure}

Direct training is a promising approach for high-performance and efficient deep SNNs.
It can be mainly divided into unsupervised and supervised learning; which are represented by STDP and stochastic gradient descent (SGD), respectively.
STDP is a biologically plausible training algorithm that considers the causal relationship between the spikes of pre- and post-synaptic neurons~\cite{diehl2015unsupervised}.
While it takes into account the characteristics of SNNs, its low training performance compared to other algorithms has limited its application in deep SNN training.


%
The gradient-based training algorithm of deep SNNs leverages successful training algorithms from DNNs, such as SGD and error backpropagation.
One of the significant obstacles to training deep SNNs with a gradient-based algorithm was the non-differentiability of spiking neurons as depicted in Eq.~\ref{eq:fire}.
To overcome this, STBP with a surrogate gradient, which approximates the gradient, was proposed and could train deep SNNs successfully~\cite{wu2018spatio}.
Since then, subsequent studies on improving the training performance of deep SNNs have been published, such as threshold-dependent batch normalization (tdBN)~\cite{zheng2021going}, temporal effective batch normalization~\cite{duan2022temporal}, and temporal efficient training with time-variant target distribution~\cite{deng2022temporal}.
However, these training algorithms did not utilize local information that can improve the training performance.
Recently, a study using local information for training was published, but it did not utilize relationships between spikes~\cite{yang2022training}.



%

%

%

\section{Methods}

With the introduction of scalable training algorithms, such as STBP, deep SNNs have become trainable with gradients.
However, these existing gradient-based algorithms for deep SNNs have a limitation in that they do not effectively consider the causal relationship between spikes of pre- and post-synaptic neurons.
Thus, in this work, we propose a method to exploit the local spike information in training deep SNNs with the gradient-based algorithm.

\setlength{\textfloatsep}{0pt}
\ctable[
pos = t,
center,
caption = {Accuracy and spikes on various configurations with CIFAR10 (training results of four times repetitions)},
label = {tab:result_cifar10},
]{l|c|cccc}{
}{
    \toprule
    \multirow{2}{*}{Methods} & \multirow{2}{*}{Reset} & \multicolumn{2}{c|}{Accuracy (\%)} & \multicolumn{2}{c}{\# of Spikes (K)} \\
    & & Mean & \multicolumn{1}{c|}{Max} & Mean & \multicolumn{1}{c}{Max} \\
    \midrule
    \midrule
    \multicolumn{6}{l}{ResNet20} \\
    \midrule
    Baseline & soft & 94.91 & 94.97 & 497 & 520 \\
    Gradient scale & soft & \textbf{95.05} & \textbf{95.11} & \textbf{492} & \textbf{506} \\
    \midrule
    Baseline & hard & \textbf{93.68} & \textbf{93.86} & 463 & \textbf{470} \\
    Gradient scale & hard & 93.58 & 93.69 & \textbf{451} & 475 \\
    \midrule
    \midrule
    \multicolumn{6}{l}{ResNet32} \\
    \midrule
    Baseline & soft & 95.00 & 95.16 & 814 & 879 \\
    Gradient scale & soft & \textbf{95.12}  & \textbf{95.22} & \textbf{783} & \textbf{822} \\
    \midrule
    Baseline & hard & 90.52 & 90.71  & \textbf{555} & 572 \\
    Gradient scale & hard & \textbf{90.57} & \textbf{90.91} & 563 & \textbf{571} \\
    \bottomrule
}

\setlength{\textfloatsep}{0pt}
\ctable[
pos = t,
center,
caption = {Accuracy and spikes on various configurations with CIFAR100 (training results of four times repetitions)},
label = {tab:result_cifar100},
]{l|c|cccc}{
}{
    \toprule
    \multirow{2}{*}{Methods} & \multirow{2}{*}{Reset} & \multicolumn{2}{c|}{Accuracy (\%)} & \multicolumn{2}{c}{\# of Spikes (K)} \\
    & & Mean & \multicolumn{1}{c|}{Max} & Mean & \multicolumn{1}{c}{Max} \\
    \midrule
    \midrule
    \multicolumn{6}{l}{ResNet20} \\
    \midrule
    Baseline & soft & 74.83 & 75.18 & 641 & 649 \\
    Gradient scale & soft & \textbf{75.20} & \textbf{75.88} & \textbf{634} & \textbf{642} \\
    \midrule
    Baseline & hard & 72.26 & 72.37 & 562 & 573 \\
    Gradient scale & hard & \textbf{72.29} & \textbf{72.42} & \textbf{542} & \textbf{555} \\
    \midrule
    \midrule
    \multicolumn{6}{l}{ResNet32} \\
    \midrule
    Baseline & soft & \textbf{75.57} & \textbf{75.79} & 959 & 984 \\
    Gradient scale & soft & 75.45 & 75.73 & \textbf{954} & \textbf{979} \\
    \midrule
    Baseline & hard & 61.96 & 62.12 & 769 & \textbf{782} \\
    Gradient scale & hard & \textbf{62.03} & \textbf{62.19} & \textbf{759} & 788 \\
    
    \bottomrule
}

Before explaining the proposed method, we should define the relational expression of spikes.
There are various representations for spike relation, but, in this work, we adopt trace-based representation for its low computational complexity, which is suitable for deep SNNs~\cite{morrison2008phenomenological}.
An example of the representation is shown in Fig.~\ref{fig:spike_trace}.
$S_{\textrm{pre}}$ and $S_{\textrm{post}}$ indicate pre- and post-synaptic spike trains, respectively.
The history of spike generation in each neuron is recorded in the spike trace $x$ as follows:
\begin{equation}
x_{i}^{l}(t) = e^{\textrm{-}1} x_{i}^{l}(t\textrm{-}1) + s_{i}^{l}(t) \textrm{.}
\end{equation}
The spike trace increases by a spike when the neuron fires and exponentially decreases at each time step of the forward (blue dotted line in Fig.~\ref{fig:spike_trace})
Each layer has pre- and post-synaptic traces ($X_{\textrm{pre}}^{l}$, $X_{\textrm{post}}^{l}$) according to its connection.
With these two spike traces, we defined the relationship of spikes $R$ as
\begin{equation}
R^{l}(t) = f^{l}(X_{\textrm{pre}}^{l}(t),X_{\textrm{post}}^{l}(t)) \textrm{,}
\end{equation}
where $f^{l}$ is a relationship function of $l$th layer.
During training, it is calculated in the backward path (orange dotted line in Fig.~\ref{fig:spike_trace}).
We used convolution and outer product operations for the relationship function $f$ of convolution and fully connected layers, respectively.

\setlength{\textfloatsep}{0pt}
\ctable[
pos = t,
star,
center,
caption = {Comparisons with previous methods on CIFAR10 (KD: Knowledge Distillation)},
label = {tab:comparison_cifar10},
]{lccccccc}{
    \vspace{-6.0em}
}{
    \toprule
    \multirow{2}{*}{Methods} & \multirow{2}{*}{Training} & \multirow{2}{*}{Architecture} & \multirow{2}{*}{Neuron} & \multirow{2}{*}{Reset} & \multirow{2}{*}{T} & ANN & SNN \\
    & & & & & & Acc. (\%) & Acc. (\%)\\
    \midrule

    Calibration~\cite{li2021free} & conversion & ResNet20 & IF & soft & 32 & 95.46 & 94.78 \\
    SNN-aware training~\cite{bu2023optimal} & conversion & ResNet18 & IF & soft & 4 & 96.04 & 90.43 \\
    tdBN~\cite{zheng2021going} & direct & ResNet19 & LIF & hard & 4 & - & 92.92 \\
    TET~\cite{deng2022temporal} & direct & ResNet19 & LIF & hard & 4 & - & 94.44 \\
    Local tandem~\cite{yang2022training} & KD & ResNet20 & LIF & soft & 16 & 95.36 & 94.76 \\ 
    Gradient scale (Ours) & direct & ResNet20 & LIF & soft & 4 & - & 95.11 \\
    Gradient scale (Ours) & direct & ResNet20 & LIF & hard & 4 & - & 93.69 \\
    
    \bottomrule
    \vspace{-1.5em}
}

We proposed a gradient scale that adjusts the gradient of synaptic weight according to the local relationship of the spike.
Inspired by STDP, the proposed algorithm encourages training with the gradient when there is a causal relationship between pre- and post-synaptic spikes.
Otherwise, if there is less relationship, the algorithm hinders the training.
We implemented this encouragement and hindrance by scaling the gradients of synaptic weights obtained from STBP as follows:
%
\begin{equation} \label{eq:weight_update}
\Delta W^{l} = -\eta g(\frac{\delta L}{\delta W^{l}},R^{l}) 
    = -\eta (\alpha \frac{\delta L}{\delta W^{l}} \circ R^{l} + (1-\alpha) \frac{\delta L}{\delta W^{l}}) \textrm{,}
\end{equation}
where $L$ is a loss, $\eta$ is a learning rate, $g$ is a gradient scaling function, $\alpha$ is a interpolation coefficient, and $\circ$ is element-wise multiplication (Hadamard product).
The gradient scaling function $g$ receives the gradients and spike relationship as inputs.
As described in Eq.~\ref{eq:weight_update}, we adopted a simple linear interpolation function for the scaling.
In this work, we set the coefficient $\alpha$ to 0.1 empirically.

\section{Experiments}

\subsection{Experimental Setup}
To evaluate the effectiveness of the proposed gradient scaling, we set STBP~\cite{wu2018spatio} and tdBN~\cite{zheng2021going} as a baseline training algorithm.
For each configuration, we trained deep SNN models for 300 epochs using SGD.
We adopted a learning rate schedule in which the learning rate decreased to 0.1 times every 100 epochs.
We used LIF neurons with the leak constant $\tau$ of 0.9, and the time step was fixed to four.
We constructed deep SNN models based on ResNet20 and ResNet32 architectures and trained them on image classification datasets, such as CIFAR10 and CIFAR100.
For data augmentation, Cutmix~\cite{yun2019cutmix} was used, and for input encoding, real value encoding was applied as in other studies~\cite{wu2018spatio,zheng2021going}.


\subsection{Experimental Results}
The experimental results on CIFAR10 and CIFAR100 are presented in Tables~\ref{tab:result_cifar10} and~\ref{tab:result_cifar100}, respectively.
We compared the training results of the baseline and proposed methods in various configurations of the model architecture and reset method of spiking neurons.
For fair and precise evaluations, we recorded the mean and maximum results for test accuracy and spike count after training four times on each configuration.
For the accuracy on CIFAR10 dataset, the mean and maximum accuracy were improved when the proposed gradient scale was applied in all cases except for the case of ResNet20 with the hard reset, as shown in Table~\ref{tab:result_cifar10}.
Furthermore, the proposed approach can reduce the number of spikes in most cases.
There are similar trends in training results on CIFAR100, as depicted in Table~\ref{tab:result_cifar100}.
The accuracy and spike counts are improved in most cases except ResNet32 with the soft reset and the hard reset, respectively, with the proposed methods.

Table 3 presents the comparisons of the proposed method with other deep SNN training methods.
For a fair comparison, we compared the results of a model structure similar to ResNet20 on the CIFAR10 dataset.
Overall, the proposed approach shows higher training performance than the recent previous methods.
In the case of soft reset, when the proposed method is applied, we achieve higher accuracy with shorter time steps than the conversion methods~\cite{li2021free,bu2023optimal} and local tandem learning~\cite{yang2022training}.
In the case of hard reset, it shows higher accuracy than tdBN~\cite{zheng2021going}, but lower training performance than TET~\cite{deng2022temporal}.
It was difficult to compare other metrics, such as spike counts, in this study as they were not commonly reported in previous works.

\section{Discussion}

The proposed training algorithm can be further improved with optimization of relation function $f$, scaling function $g$, and hyperparameters, such as $\alpha$ in Eq.~\ref{eq:weight_update}.
In this paper, a simple spike relation function, which only considers the positive relation between the spike traces, and scaling function was used to show the feasibility of enhancement training performance with local spike information.
In order to the improvement, we can consider more complicated relation functions of pre- and post-synaptic spike traces, which consider a negative relation of the spike traces as the STDP learning rule.
Furthermore, we can use other scaling functions based on theoretical analysis of deep SNNs, instead of linear interpolation as in this work.

%
%
%
%

\section{Conclusion}

In this paper, we proposed a training method for deep SNNs with spike-dependent local information.
The proposed method, which is compatible with gradient-based training algorithms, such as STBP, scales the gradient of synaptic weight according to the relationship between spike traces of adjacent neurons.
We verified the effectiveness of the proposed approach with ResNet architecture on CIFAR datasets.
In the future, we will improve the proposed algorithm through exploration and optimization of the spike relation function and gradient scaling function.
In addition, we will evaluate the algorithm with other model architectures and datasets.
We believe that by taking into account the characteristics of SNNs and utilizing local information, the training performance of deep SNNs can be improved.

\section*{Acknowledgements}

This work was supported in part by the Korea Institute of Science and Technology (KIST) through 2E32260 and the National Research Foundation of Korea (NRF) grant funded by the Korea government (Ministry of Science and ICT) [NRF-2021R1C1C2010454].

\nocite{langley00}

\bibliography{main}
\bibliographystyle{icml2023}

%

\end{document}